\documentclass[runningheads,a4paper]{llncs}

\usepackage{amssymb}
\usepackage{graphicx}
\usepackage{floatrow}
\usepackage[colorlinks=true,citecolor=blue,urlcolor=black]{hyperref}
\newfloatcommand{capbtabbox}{table}[][\FBwidth]
\usepackage{blindtext}
\usepackage{pifont}
\newcommand{\cmark}{\ding{51}}%
\newcommand{\xmark}{\ding{55}}%

\begin{document}

\mainmatter  

\title{Small Organ Segmentation in Whole-body MRI using a Two-stage FCN and Weighting Schemes}

\titlerunning{Small Organ Segmentation in Whole-body MRI}

\author{Vanya V. Valindria$^1$ \and Ioannis Lavdas$^2$ \and  Juan Cerrolaza$^1$ \and Eric O. Aboagye$^2$ \and\\Andrea G. Rockall$^{2}$ \and Daniel Rueckert$^1$ \and Ben Glocker$^1$}
\institute{$^1$Biomedical Image Analysis Group, Department of Computing\\
$^2$Comprehensive Cancer Imaging Centre, Department of Surgery and Cancer\\ 
Imperial College London, UK}

%
\authorrunning{Valindria et al.}


%
%

\toctitle{Small Organ Segmentation}
\tocauthor{Vanya Vabrina Valindria}
\maketitle

\begin{abstract}

Accurate and robust segmentation of small organs in whole-body MRI is difficult due to anatomical variation and class imbalance. Recent deep network based approaches have demonstrated promising performance on abdominal multi-organ segmentations. However, the performance on small organs is still suboptimal as these occupy only small regions of the whole-body volumes with unclear boundaries and variable shapes. A coarse-to-fine, hierarchical strategy is a common approach to alleviate this problem, however, this might miss useful contextual information. We propose a two-stage approach with weighting schemes based on auto-context and spatial atlas priors. Our experiments show that the proposed approach can boost the segmentation accuracy of multiple small organs in whole-body MRI scans.
\end{abstract}

\section{Introduction}
Multi-organ segmentation in abdominal and whole-body scans is challenging as there are various organs and structures the need to be captured simultaneously. The size, shape and appearance of abdominal organs vary considerably between patients, but also the relative positions change to some degree. Small organs are less often investigated compared to major organs, although small organs are of interest for diagnosis and clinical applications such as cancer screening. Machine learning methods have been used to segment multiple organs in abdominal images \cite{lavdas2017fully}. However, small organs are still underrepresented and show lower accuracies compared to the large ones with less shape variability (e.g., lungs, heart, spine). Small object segmentation is generally more challenging due to large class imbalance between object and background samples. For example, the ratio of small organs in whole-body MRI in our data is less than 0.007\% of the overall volume. We focus on bladder, sacrum, rectum, clavicles, pancreas, gall-bladder, and adrenal gland. The complexity of background intensity and weak boundaries often make it more difficult to segment small organs.

Most of the multi-organ segmentation work is applied to CT for which data seems more widely available. We focus on MRI, as there are still fewer works on this modality while whole-body MRI has become an important diagnostic tool for cancer screening. Early works of multi-organ abdominal segmentation include multi-atlas label fusion \cite{bai2013probabilistic} and statistical shape models \cite{cerrolaza2016soft}. Multi-atlas techniques register images from a reference database to each new image and fuse multiple atlases to obtain the final segmentation.  More recent work has made use of deep learning, for example, architectures for multi-organ segmentation, such as the dense V-network \cite{gibson2018automatic}. Weighted U-Net with weight proportion for foreground and background has been used to address the class imbalance problem \cite{ronneberger2015u}. 

In the medical domain, a coarse-to-fine approach has been applied for small structures, such as lesion segmentation in the pancreas \cite{zhou2017deep} and liver \cite{christ2016automatic}. Combination of multi-atlas and CNN techniques were used in \cite{larsson2017robust}, where localization of region of interest using a multi-atlas approach is combined with voxel-wise binary classification using CNNs. A more advanced iterative coarse-to-fine approach has been proposed by \cite{zhou2017fixed}, which uses a smaller input region for a more accurate segmentation from multi-view coarse segmentations. However, finer segmentation consists of iterative refinement of at least 10 iterations, which can be time consuming. 

Pancreas is the most studied small organ in previous works, as it is an abdominal organ of great importance with high anatomical variability. An earlier work on pancreas segmentation by \cite{roth2015deeporgan} combines regional CNNs with superpixels at multiple scales. Later, \cite{roth2016spatial} integrates semantic mid-level cues (organ interior and boundary maps) via spatial aggregation and \cite{cai2017improving} applied long short-term memory (LSTM), to address the contextual learning and pancreas segmentation consistency problem. A novel approach with self-attention gating in CNNs to segment the pancreas is introduced in \cite{oktay2018attention} for a more specific local region segmentation. As an end-to-end approach, they showed an improvement in pancreas segmentation, compared to standard FCNs, dense dilated FCNs \cite{gibson2018automatic}, holistically nested FCNs \cite{roth2016spatial}, and standard U-Net. To alleviate the missing contextual information in the common two-stage approach, a recurrent saliency transformation network was proposed to relate the coarse and fine stages \cite{yu2018recurrent}. This saliency transformation module repeatedly transforms the segmentation probability map from previous iterations as spatial priors. However, performance can be lower than the coarse segmentation results because of the unsatisfying convergence over iterations.

Our contributions are as follows: (1) Previous work focuses on CT segmentation and a single organ \cite{cai2017improving,roth2015deeporgan,roth2016spatial,gibson2018automatic,oktay2018attention}, while we study segmentation of multiple small organs on whole-body MRI including structures such as bones rarely considered. (2) This work is based on a coarse-to-fine framework \cite{cai2017improving,yu2015multi,zhou2017fixed} but goes one step further by incorporating weighting schemes and a specialized ROI selection. Weighting helps with class imbalance. For fine-scale segmentation, we apply auto-context with spatial information obtained from atlases so that coarse-and-fine-scaled networks are optimized jointly.

\section{Materials and Methods}
\subsection{Materials}

In-house whole-body MRI data was obtained from 48 healthy volunteers using the protocol described in \cite{lavdas2015apparent}. Segmentations include 11 abdominal organs (heart, right lung, left lung, liver, adrenal gland, gall bladder, right kidney, left kidney, spleen, pancreas, and bladder) and 7 bones (spine, right clavicle, left clavicle, pelvis, humerus, sacrum). For our experiments, we only use T2w sequences, which we were resampled to a size (112, 80, 256) with isotropic 4 mm spacing.

\subsection{Two-stage network: A coarse-to-fine approach}

Our aim is to segment multiple small organs from MRI scans, which occupy only a very small part of an MRI volume. We apply a two-stage network, which has been shown to be successful in an organ segmentation task \cite{cai2017improving,zhou2017fixed,roth2015deeporgan}. CNN-based methods produce less accurate results when detecting small organs, particularly because the network is confused by the complicated context in the background and other organs. A coarse-scale is first used to locate the organ of interest for a subsequent fine-scale organ segmentation.

\begin{figure}
\centering
\includegraphics[width=0.9\linewidth]{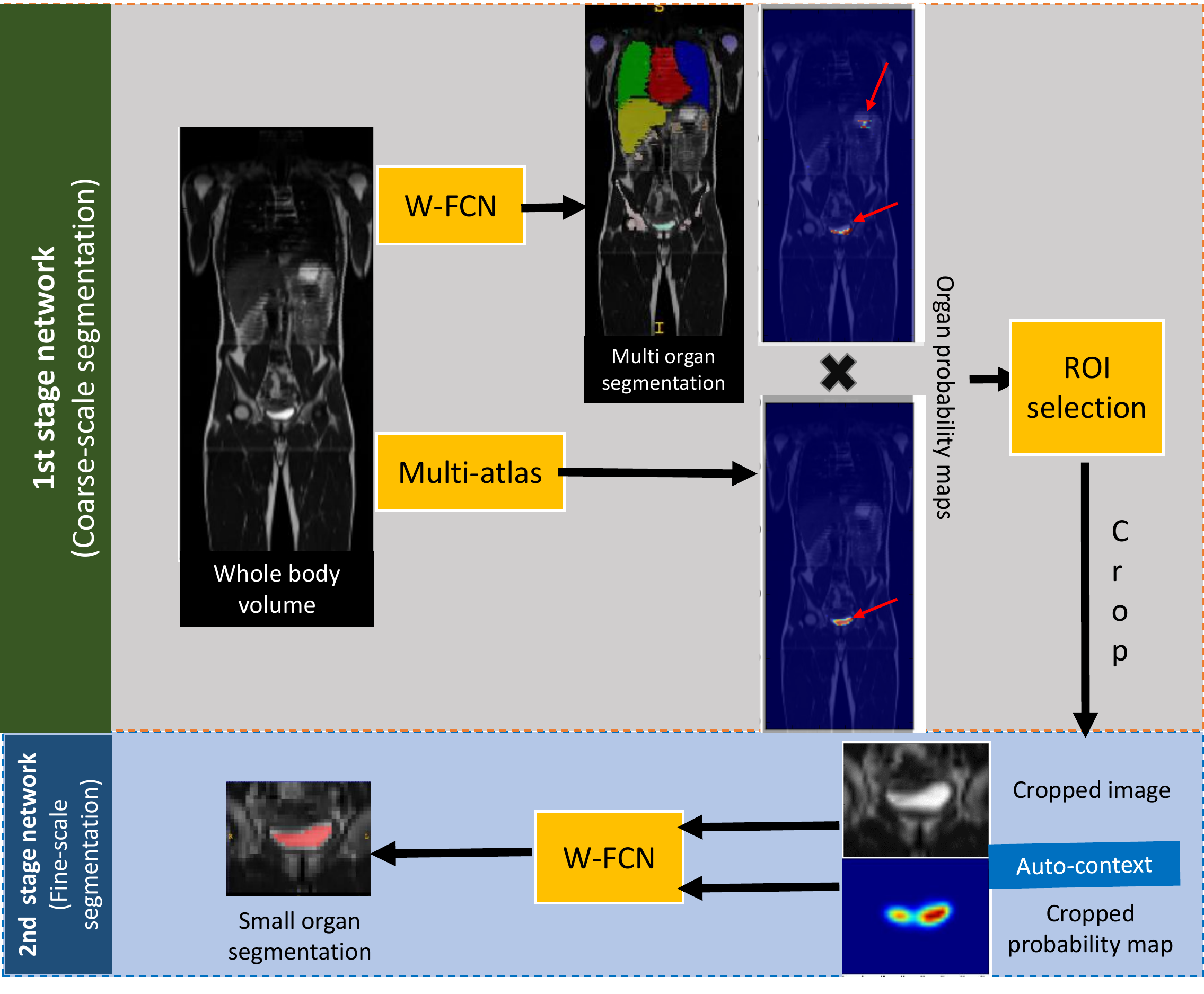}
\caption{Overview. First stage: Coarse-scale segmentation with multi-organ segmentation with weighted-FCN, where we obtain the segmentation results and probability map for each organ. Second stage: Fine-scale binary segmentation per organ. The input consists of a cropped volume and a probability map from coarse segmentation.}
\label{fig:overview}
\end{figure}

The two-stage strategy effectively reduces the complexity of the background while enhancing the discriminative information of a small organ. We train a coarse-scale segmentation to deal with the multi-organ segmentation on the whole-body scan. Then, at the fine-scale segmentation, we only focus with the ROI selection according to the coarse segmentation. On each stage, two different networks were trained respectively. Similarly in testing stage, the coarse-scale network was first used to obtain the rough position of multiple small organs. Then, fine-scale networks were employed for binary segmentation for each of the small organs.

To counter the class imbalance of the small organs, we use a weighted-FCN \cite{ronneberger2015u} in both segmentation scales (coarse and fine). For coarse segmentation we use a weighted-FCN for multi-class segmentation with different weights set for each class. Small weights were set for bigger organs (such as livers, lungs, etc.) and large weights were applied for small organs, according to the relative size to the whole body volumes. This per-class-weight proportion was chosen via experimentation and the statistics of each organ proportion in the database.

\subsubsection{ROI Selection} \label{roi_chapter}
In whole-body scans, each organ is located in a specific region. Therefore, context and spatial information is crucial for organ segmentation. After organ localisation from the coarse-scale segmentation, we found that in some cases, the FCN fails to locate small organs, leading to a much wider ROI for the fine-scale segmentation. Therefore, we need to induce spatial priors in order to guide the second stage network with a better ROI. A multi-atlas technique was chosen for producing spatial priors, because with multi-atlas segmentation, although the segmentation accuracy is lower than CNN-based accuracy, it mostly generates good organ localisation \cite{lavdas2017fully}. As shown in Fig.\ref{fig:overview}, the bladder segmentation result of multi-atlas is focused in one specific location instead of having multiple predictions scattered in other areas (shown in FCN probability maps).

In the multi-atlas approach, each atlas image is registered to the target and then fused using the methods described in \cite{bai2013probabilistic}. We chose the PBAF (patch-based segmentation with augmented features) label fusion technique, as compared to other methods, this provides better accuracy for small organs. While \cite{larsson2017robust} used a multi-atlas approach for organ localisation at first step before binary segmentation via CNNs, we only incorporate it after the coarse-scale segmentation. Here, the probability maps of coarse-segmentation is multiplied with the one from the multi-atlas approach (as a spatial prior for false-positive reduction, so that the fidelity of the final probability map is improved) to produce the 3D bounding box for fine-scale segmentation, see Fig.\ref{fig:overview} top for details. We denoted this part of spatial prior incorporation as combined multi-atlas.

\subsubsection{Auto-context}
We utilize the classical approach of auto-context \cite{tu2010auto} into our framework to fuse and to integrate the information from different stages with the context. The problem of coarse-to-fine segmentation is that sometimes the cropped ROI within the bounding box has less sufficient spatial context, making the fine-scale networks more confused than the coarse-scale segmentation \cite{roth2016spatial}. Hence, we incorporate the probability maps from the coarse-scale segmentation into the fine-scale segmentation. The benefit of probability maps as visual cues have been discussed in \cite{roth2016spatial} as a spatial aggregation from multi-view segmentation and in \cite{yu2018recurrent} as an updated input for an iterative fine-scale segmentation with saliency transformation network. In our case, a simple auto-context with probability maps incorporation helps the FCN to integrate the information from coarse-to-fine level segmentation. By using additional input (see Fig.\ref{fig:overview} - bottom), uncertainties and errors from the first stage are adjusted.

\section{Experiments and Results}

We use different strategies to achieve small organ segmentation. For coarse and fine scale segmentation, we use the same architecture of an FCN with residual layers, as implemented in \cite{pawlowski2017DLTK}. We train the network using Adam optimisation with a learning rate of 0.001, $\beta_1 = 0.9 $ and $\beta_2 = 0.999$,  $\epsilon = 10^{-5} $. In coarse-scale segmentation, we use mini-batch training of 16 training examples with size $64^3$, to provide enough context at 4 mm resolution, while in fine-scale segmentation, the example size are $8^3$ to fit with the input images. We train each networks with 10K iterations. 

We perform two-fold cross-validation, where we split the dataset into two fixed folds with equal number of samples. For evaluation, we measure the segmentation accuracy by computing the Dice Similarity Coefficient (DSC) for each subject.

\begin{table}[]
\centering
\caption{Different strategies on segmentation of small organs, using the baseline FCN, weighted-FCN (W), auto-context (AC), two-stage networks (2SN), and combined multi-atlas (CMA) for ROI selection. Improvements in overall small organ segmentation accuracy (reported in DSCs) was achieved with our proposed approach.}
\label{tab:fine_segm}
\resizebox{\columnwidth}{!}{%
\begin{tabular}{ccccccccccccc}
\hline
\multicolumn{4}{c}{Approaches}               & \multicolumn{1}{c}{Adrenal Gland} & \multicolumn{1}{c}{Gall Bladder} & \multicolumn{1}{c}{Pancreas} & \multicolumn{1}{c}{Bladder} & \multicolumn{1}{c}{R. Clavicle} & \multicolumn{1}{c}{L. Clavicle} & \multicolumn{1}{c}{Rectum} & \multicolumn{1}{c}{Sacrum} & \multicolumn{1}{c}{Avg.} \\ \hline
\multicolumn{4}{c}{Multi-atlas} & 0.044                             & 0.298                            & 0.377                        & 0.680                       & 0.135                           & 0.163                           & 0.429                      & 0.466                      & 0.324                    \\ \hline
\multicolumn{4}{c}{FCN}                      &                                   &                                  &                              &                             &                                 &                                 &                            &                            &                          \\ \cline{1-4}
W        & AC        & 2SN       & CMA       &                                   &                                  &                              &                             &                                 &                                 &                            &                            &                          \\ \hline
\xmark        & \xmark         & \xmark         & \xmark         & 0.006                             & 0.330                            & 0.373                        & 0.401                       & 0.055                           & 0.080                           & 0.346                      & 0.361                      & 0.244                    \\ \hline
\cmark        & \xmark         & \xmark         & \xmark         & 0.097                             & 0.443                            & 0.465                        & 0.689                       & 0.462                           & 0.484                           & 0.526                      & 0.701                      & 0.483                    \\ \hline
\cmark        & \xmark         & \xmark         & \cmark         & \textbf{0.203}                    & 0.436                            & 0.519                        & 0.717                       & 0.381                           & 0.397                           & 0.549                      & 0.693                      & 0.487                    \\ \hline
\xmark        & \xmark         & \cmark         & \xmark         & 0.046                             & 0.375                            & 0.455                        & 0.594                       & 0.274                           & 0.379                           & 0.393                      & 0.490                      & 0.376                    \\ \hline
\cmark        & \cmark        & \cmark         & \xmark         & 0.133                             & 0.507                            & \textbf{0.612}               & 0.729                       & 0.519                           & 0.535                           & 0.591                      & 0.721                      & 0.543                    \\ \hline
\cmark        & \cmark         & \cmark         & \cmark         & 0.146                             & \textbf{0.532}                   & 0.567                        & \textbf{0.754}              & \textbf{0.541}                  & \textbf{0.547}                  & \textbf{0.658}             & \textbf{0.735}             & \textbf{0.560}           \\ \hline
\end{tabular}%
}
\end{table}

First stage of the network was trained for multi-organ (18 classes, including large organs) on the entire 3D whole-body ROI. However, for our study, we were only interested in small organs as large organs, such as heart, liver, lungs, and spine - have achieved satisfactory results (with above 0.9 DSC). Our complete benchmarks for different strategies applied in small organ segmentation are shown in Table \ref{tab:fine_segm}. Although the overall results on 18 organs for multi-atlas segmentation are lower (DSC: 0.486) than baseline FCN (DSC: 0.516) (without weights, auto-context, two-stage network and combined multi-atlas), the accuracy for small organs improved. 

We then investigate the role of introducing different weights when taking training samples from different classes on coarse-scale multi-organ segmentation. As we give larger weights on small organs, we can reduce the effect of class imbalance \cite{ronneberger2015u}. This weighted-FCN gives significant improvement over small organs compared to the baseline FCN, with overall small organ accuracies increase from 0.244 to 0.483 DSC.

To evaluate the spatial prior in combined multi-atlas, we multiply the probability maps from the weighted-FCN with the multi-atlas prediction. Table \ref{tab:fine_segm} shows that it slightly improves the accuracy for small organ segmentation, especially adrenal gland (DSC: 0.203). However, some organs, such as both clavicles are worse because of the lower accuracy on multi-atlas segmentation. The probability maps given by multi-atlas are tighter, so that the segmentation misses some parts of the region of interested.

To verify that the two-stage network can segment small organs more accurately, we run the state-of-the-art method \cite{zhou2017deep}, which takes input from baseline FCN segmentation to crop the ROI for fine-scale segmentation. Compared to the small organ segmentation accuracies produced by the FCN baseline (DSC: 0.244), we observe an improvement of about 50\%. This result shows the advantage of using a two-stage network on small organ segmentation.

We then employ a two-stage scheme using the weighted-FCN with auto-context and the probability map from the weighted-FCN prediction as an additional input to the network. With this strategy, we get much higher accuracy on all organs (see Table \ref{tab:fine_segm}). Weightings, auto-context, and region cropping are shown to boost the performance of small organ segmentation. 

To add the spatial prior, we crop the ROI for fine-scale segmentation for organ-specific bounding box, according to the Section~\ref{roi_chapter}. As detailed in Table~\ref{tab:fine_segm}, adding spatial priors is useful in almost all small organs, except the pancreas because of its poor tissue contrast and shape variability. For adrenal gland, too, we find that the results of direct multiplication between coarse-scale segmentation and spatial prior gives better results. Multi-atlas prior information gives better localisation for adrenal gland, which is the smallest organ in our task (only occupied about 0.0001\% of the whole volume). Overall, the addition of auto-context and spatial priors to two-stage weighted-FCN gives the best results on small organ segmentation (DSC: 0.560), as detailed in Table~\ref{tab:fine_segm}.


\begin{figure}
\centering
\includegraphics[width=0.49\linewidth]{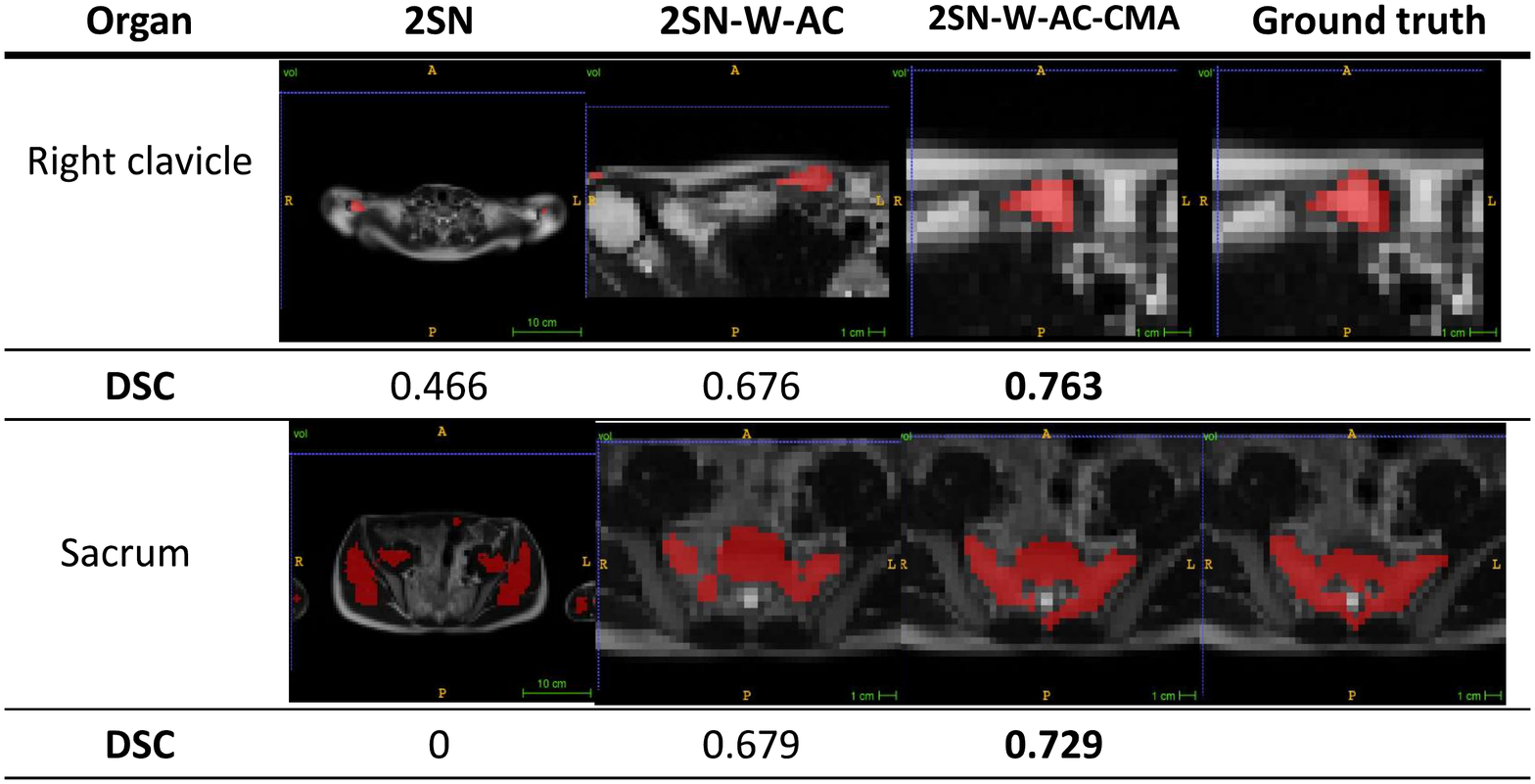}
\includegraphics[width=0.49\linewidth]{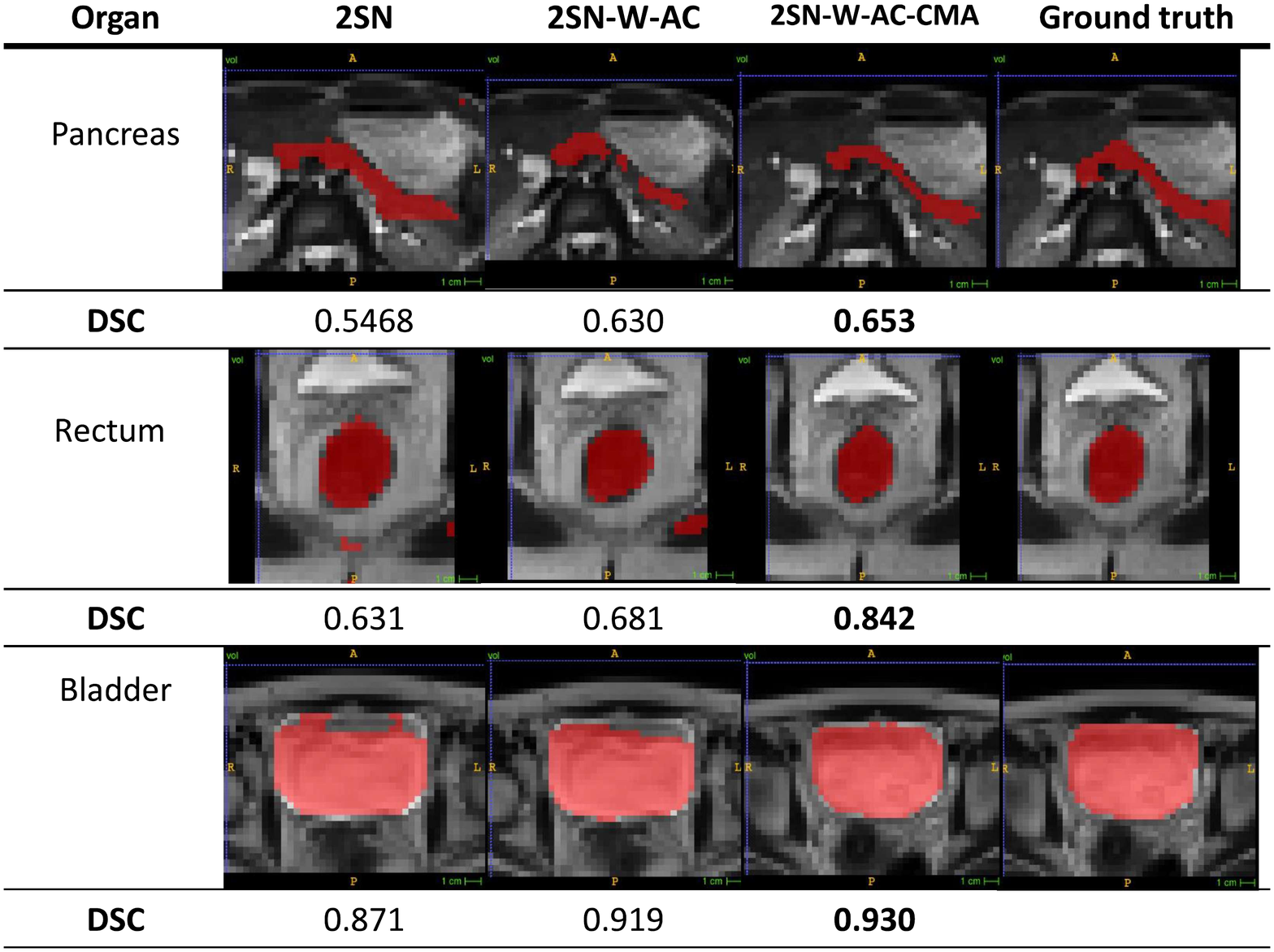}
\caption{Fine-scale segmentation of small organs. Two-stage network (2SN) with weighting (W), auto-context (AC), and combined spatial prior from multi-atlas (CMA) can improve small organ segmentation results.}
\label{fig:fine_organ1}
\end{figure}

\section{Discussion and Conclusion}

The challenge in small organ segmentation is our main motivation in this work. A standard CNN-based approach gives good performance on larger organs, but still fails to accurately segment small organs. We found that by setting different weights on training examples per class could boost the performance of small organ segmentation. The class imbalance problem is countered by a multi-class weighted-FCN. As the target is often very small, we need to focus on a local input region. The two-stage network scheme seems to help the small organ segmentation. However, lack of contextual information and spatial knowledge sometimes make the network confused. Hence, we apply a simple auto-context, which uses the coarse-scale probability map to carry useful context information for fine-scale segmentation. 

Some examples of small organ fine-scale segmentation are shown in Figure~ \ref{fig:fine_organ1}. We can see that the two-stage weighted-FCN with auto-context and combination with multi-atlas produce better results in small organ segmentation. False positives are reduced and the ROIs are more focused to the specific organ. Our experiments show that the proposed approach outperforms the baseline (standard FCN and previous state-of-the art) on multiple small organs segmentations. This work shows a promising result for small organ and bone segmentation in whole-body MRI. 

\subsection*{Acknowledgments}
\small{
V. Valindria is supported by the Indonesia Endowment for Education (LPDP)- Indonesian Presidential PhD Scholarship programme. B. Glocker received funding from the European Research Council (ERC) under the EU's Horizon 2020 research and innovation programme (grant agreement No 757173, project MIRA, ERC-2017-STG).

The MRI data has been collected as part of the MALIBO project funded by the Efficacy and Mechanism Evaluation (EME) Programme, an MRC and NIHR partnership (EME project 13/122/01). The views expressed in this publication are those of the authors and not necessarily those of the MRC, NHS, NIHR or the Department of Health.}

\bibliography{ref_DA}

\begin{thebibliography}{10}
\providecommand{\url}[1]{\texttt{#1}}
\providecommand{\urlprefix}{URL }

\bibitem{bai2013probabilistic}
Bai, W., Shi, W., O'Regan, D.P., Tong, T., Wang, H., Jamil-Copley, S., Peters,
  N.S., Rueckert, D.: A probabilistic patch-based label fusion model for
  multi-atlas segmentation with registration refinement: application to cardiac
  {MR} images. IEEE transactions on medical imaging  32(7),  1302--1315 (2013)

\bibitem{cai2017improving}
Cai, J., Lu, L., Xie, Y., Xing, F., Yang, L.: Improving deep pancreas
  segmentation in {CT} and {MRI} images via recurrent neural contextual
  learning and direct loss function. arXiv preprint arXiv:1707.04912  (2017)

\bibitem{cerrolaza2016soft}
Cerrolaza, J.J., Summers, R.M., Linguraru, M.G.: Soft multi-organ shape models
  via generalized {PCA}: A general framework. In: MICCAI. Springer (2016)

\bibitem{christ2016automatic}
Christ, P.F., Elshaer, M.E.A., Ettlinger, F., Tatavarty, S., Bickel, M., Bilic,
  P., Rempfler, M., Armbruster, M., Hofmann, F., D’Anastasi, M., et~al.:
  Automatic liver and lesion segmentation in {CT} using cascaded fully
  convolutional neural networks and {3D} conditional random fields. In:
  International Conference on Medical Image Computing and Computer-Assisted
  Intervention. pp. 415--423. Springer (2016)

\bibitem{gibson2018automatic}
Gibson, E., Giganti, F., Hu, Y., Bonmati, E., Bandula, S., Gurusamy, K.,
  Davidson, B., Pereira, S.P., Clarkson, M.J., Barratt, D.C.: Automatic
  multi-organ segmentation on abdominal {CT} with dense {V-networks}. IEEE
  Transactions on Medical Imaging  (2018)

\bibitem{larsson2017robust}
Larsson, M., Zhang, Y., Kahl, F.: Robust abdominal organ segmentation using
  regional {CNNs}. In: Scandinavian Conference on Image Analysis. pp. 41--52
  (2017)

\bibitem{lavdas2017fully}
Lavdas, I., Glocker, B., Kamnitsas, K., Rueckert, D., Mair, H., Sandhu, A.,
  Taylor, S.A., Aboagye, E.O., Rockall, A.G.: Fully automatic, multiorgan
  segmentation in normal whole body magnetic resonance imaging {(MRI)}, using
  classification forests {(CFs)}, convolutional neural networks {(CNNs)}, and a
  multi-atlas {(MA)} approach. Medical physics  44(10),  5210--5220 (2017)

\bibitem{lavdas2015apparent}
Lavdas, I., Rockall, A.G., Castelli, F., Sandhu, R.S., Papadaki, A.,
  Honeyfield, L., Waldman, A.D., Aboagye, E.O.: Apparent diffusion coefficient
  of normal abdominal organs and bone marrow from whole-body {DWI} at 1.5 {T}:
  the effect of sex and age. American Journal of Roentgenology  205(2),
  242--250 (2015)

\bibitem{oktay2018attention}
Oktay, O., Schlemper, J., Folgoc, L.L., Lee, M., Heinrich, M., Misawa, K.,
  Mori, K., McDonagh, S., Hammerla, N.Y., Kainz, B., Glocker, B., Rueckert, D.:
  Attention {U}-net: Learning where to look for the pancreas. arXiv:1804.03999
  (2018)

\bibitem{pawlowski2017DLTK}
Pawlowski, N., Ktena, S.I., Lee, M.C., Kainz, B., Rueckert, D., Glocker, B.,
  Rajchl, M.: {DLTK}: State of the art reference implementations for deep
  learning on medical images. In: Medical Imaging meet NIPS Workshop (2017)

\bibitem{ronneberger2015u}
Ronneberger, O., Fischer, P., Brox, T.: {U}-net: Convolutional networks for
  biomedical image segmentation. In: MICCAI. pp. 234--241. Springer (2015)

\bibitem{roth2015deeporgan}
Roth, H.R., Lu, L., Farag, A., Shin, H.C., Liu, J., Turkbey, E.B., Summers,
  R.M.: Deeporgan: Multi-level deep convolutional networks for automated
  pancreas segmentation. In: MICCAI. pp. 556--564. Springer (2015)

\bibitem{roth2016spatial}
Roth, H.R., Lu, L., Farag, A., Sohn, A., Summers, R.M.: Spatial aggregation of
  holistically-nested networks for automated pancreas segmentation. In: MICCAI.
  pp. 451--459. Springer (2016)

\bibitem{tu2010auto}
Tu, Z., Bai, X.: Auto-context and its application to high-level vision tasks
  and {3D} brain image segmentation. IEEE PAMI  32(10),  1744--1757 (2010)

\bibitem{yu2015multi}
Yu, F., Koltun, V.: Multi-scale context aggregation by dilated convolutions.
  ICLR  (2016)

\bibitem{yu2018recurrent}
Yu, Q., Xie, L., Wang, Y., Zhou, Y., Fishman, E.K., Yuille, A.L.: Recurrent
  saliency transformation network: Incorporating multi-stage visual cues for
  small organ segmentation. CVPR  (2018)

\bibitem{zhou2017deep}
Zhou, Y., Xie, L., Fishman, E.K., Yuille, A.L.: Deep supervision for pancreatic
  cyst segmentation in abdominal {CT} scans. In: MICCAI (2017)

\bibitem{zhou2017fixed}
Zhou, Y., Xie, L., Shen, W., Wang, Y., Fishman, E.K., Yuille, A.L.: A
  fixed-point model for pancreas segmentation in abdominal {CT} scans. In:
  MICCAI (2017)

\end{thebibliography}
\bibliographystyle{splncs03}

\end{document}